\documentclass{article}

\usepackage{PRIMEarxiv}

\usepackage[utf8]{inputenc} 
\usepackage[T1]{fontenc}    
\usepackage{hyperref}       
\usepackage{url}            
\usepackage{booktabs}       
\usepackage{amsfonts}       
\usepackage{nicefrac}       
\usepackage{microtype}      
\usepackage{fancyhdr}       
\usepackage{graphicx}       
\graphicspath{{media/}}     

\title{Age Group Discrimination via Free Handwriting Indicators
}

\author{
  Eugenio Lomurno, Simone Toffoli, Davide Di Febbo, Matteo Matteucci, Francesca Lunardini, Simona Ferrante \\
  Politecnico di Milano \\
  Department of Electronics, Information and Bioengineering\\
  Via Ponzio 34/5, 20133 Milan, Italy\\
  \texttt{\{name.surname\}@polimi.it} \\
}

\begin{document}
\maketitle

\begin{abstract}
The growing global elderly population is expected to increase the prevalence of frailty, posing significant challenges to healthcare systems. Frailty, a syndrome associated with ageing, is characterised by progressive health decline, increased vulnerability to stressors and increased risk of mortality. It represents a significant burden on public health and reduces the quality of life of those affected. The lack of a universally accepted method to assess frailty and a standardised definition highlights a critical research gap.
Given this lack and the importance of early prevention, this study presents an innovative approach using an instrumented ink pen to ecologically assess handwriting for age group classification. Content-free handwriting data from 80 healthy participants in different age groups (20-40, 41-60, 61-70 and 70+) were analysed. Fourteen gesture- and tremor-related indicators were computed from the raw data and used in five classification tasks. These tasks included discriminating between adjacent and non-adjacent age groups using Catboost and Logistic Regression classifiers.
Results indicate exceptional classifier performance, with accuracy ranging from 82.5\% to 97.5\%, precision from 81.8\% to 100\%, recall from 75\% to 100\% and ROC-AUC from 92.2\% to 100\%. Model interpretability, facilitated by SHAP analysis, revealed age-dependent sensitivity of temporal and tremor-related handwriting features. Importantly, this classification method offers potential for early detection of abnormal signs of ageing in uncontrolled settings such as remote home monitoring, thereby addressing the critical issue of frailty detection and contributing to improved care for older adults.
\end{abstract}


\section{Introduction}\label{sec:introduction}
The worldwide increase in the elderly population is expected to grow the prevalence of frailty in older people~\cite{emiel-lancet}. Frailty is a clinical syndrome with a higher prevalence in older adults, defining a progressive decline, together with increased vulnerability to stressful factors and an increased risk of mortality~\cite{fried_fenotipe}. Frailty leads to hospitalisation and admission to long-term care, with a significant impact on public care systems~\cite{polidori, Ensrud} and a poor quality of life for those directly affected~\cite{Kojima, Hoogendijk}. At present, there is no universally accepted way of identifying frailty and a standardised definition of the conditions remains an open point~\cite{Dent}. 

A consistent concept in the literature is that frailty is a complex and dynamic process. Complex because it involves both physical and cognitive systems, and dynamic because individuals tend to progress towards states of increasing severity of frailty~\cite{Markle}. Early detection of frailty is therefore key to slowing and preventing the worsening of the syndrome until it reaches the severe, irreversible stage of pre-death~\cite{Puts}. However, detection of early symptoms is hampered by the similarity to normal ageing and the diversity of the phenotype~\cite{Gordon}. According to Fried and colleagues~\cite{fried_fenotipe}, frailty can be recognised when at least three of the following five signs are present: weakness, slow gait, low physical activity, fatigue and weight loss. If only one or two symptoms are present, a person could be considered pre-frail~\cite{fried_fenotipe}. The frailty index~\cite{mitnitski2001accumulation, Rockwood} is another proposed method to diagnose frailty. However, thresholds for frailty or pre-frailty are not universally agreed among practitioners.

Early intervention is the most effective solution for preventing the worst consequences of frailty. Therefore, much attention should be paid to vulnerable people aged 65 years and older who are at risk of becoming pre-frail~\cite{Trevis}. However, the scarcity of medical resources often leads to a delayed diagnosis of frailty. To avoid this risk, an emerging solution consists of remote monitoring technologies used to continuously track the health status of community-dwelling seniors~\cite{Chkeir}. To detect early signs of decline, particular attention has been paid to the monitoring of daily activities~\cite{lunardini2019movecare}. Indeed, in older adults, any variation in the performance of daily tasks may conceal meaningful information about decline~\cite{Buchman}. Among daily tasks, handwriting may be an optimal candidate for remote monitoring because it is a high-level skill that involves several cerebral and motor districts~\cite{Plamondon}. Therefore, it undergoes significant variations with physiological or pathological age-related decline~\cite{Rosenblum} and with specific aspects of the frailty phenotype~\cite{Camicioli}. Indeed, the quantitative analysis of handwriting has been observed to be sensitive to several neuro-motor disorders, including Parkinson's disease~\cite{Walton}, dystonia~\cite{Zeuner}, Huntington's disease~\cite{Caligiuri} and essential tremor~\cite{Alty}. 

The limitations of home-based handwriting monitoring lie in the devices available for data collection. Most studies in the literature have used commercially available tablets and digitising surfaces to study writing activities; however, the diffuse technological illiteracy of older adults makes their everyday use rather intrusive~\cite{Rosenblum, Camicioli, Brooks}. Furthermore, most previous research has analysed handwriting in controlled settings, i.e. using a standard writing protocol or selecting predefined writing sequences~\cite{Alty}. Instead, the home environment represents an uncontrolled context in which the results of standard tests cannot be assumed to be valid without supervision~\cite{Schmuckler}.

In a recent work of our research group~\cite{lunardini_difebbo}, we presented an instrumented ink pen for the automatic acquisition and quantitative analysis of handwriting to allow ecological home monitoring of writing activity~\cite{toffoli2022activity}. The tool can be used for everyday paper writing tasks and the data collection is fully automated. It does not require any further intervention by the user and therefore meets the requirements of ecological validity. We have previously investigated the reliability of handwriting and tremor indicators in healthy subjects of different ages. We then demonstrated the ability of handwriting and tremor indicators to discriminate age groups in semi-uncontrolled (i.e, the acquisitions were supervised by an operator, while the content was left free to the subjects) conditions using paper-and-pen free writing tasks. Correctly assigning a subject to his or her age group through free writing analysis can be a powerful tool for detecting abnormalities associated with age-related decline~\cite{Jin}. Especially for pre-frail individuals, a potential affinity of their writing parameters with those generally observed in a category of older subjects could be a sign of an amplified consequence of normal ageing and be interpreted as a prompt for further investigation.

In this work, we studied the handwriting indicators ability of~\cite{lunardini_difebbo} in the classification of four age groups of healthy subjects, performing two types of unconstrained writing tasks.
The paper is structured as follows: Section~\ref{sec:method} presents the instrument, experimental protocols, data processing and classification algorithm used in this work. Section~\ref{sec:results} presents the results and Section~\ref{sec:discussion} discusses them. Finally, Section~\ref{sec:conclusion} expresses the novelty and possible research improvements of the work.

\section{Method}\label{sec:method}
\subsection{The smart ink pen}

\begin{figure}
  \begin{center}
  \includegraphics[width=2.8in]{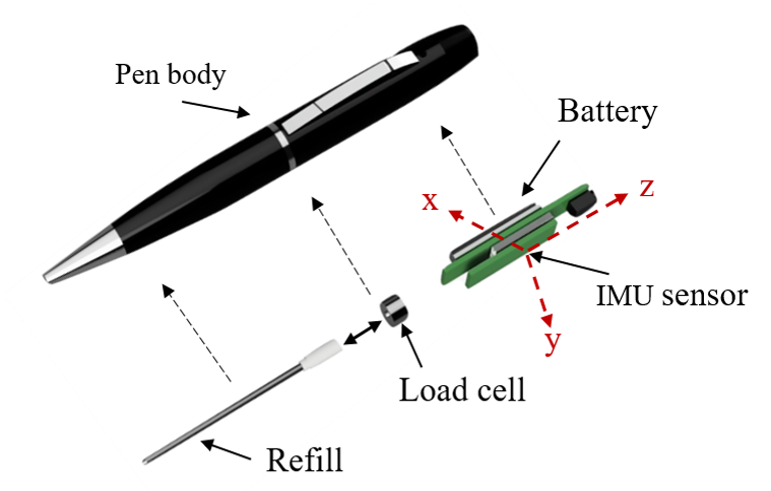}
  \caption{A digital rendering of the pen with its internal components and the IMU reference frame orientation.}\label{pen}
  \end{center}
  \vspace*{-0.6cm}
\end{figure}

We used the smart ink pen, shown in Fig.~\ref{pen}, developed in the European project MoveCare~\cite{lunardini2019movecare, FebboGNB}, to collect handwriting data. The device consists of an ink pen equipped with an inertial measurement unit (IMU) to record movement and a miniaturised load cell to record the normal force applied on the tip~\cite{FebboGNB}. The main advantage of this device is that it tests handwriting in a condition as close as possible to the normal situation, giving the typical feel of writing on paper. 

The pen is designed to automatically start collecting signals when it is moved to write, and the stored data can be accessed via Bluetooth connection. All electronic components and the data storage mechanism are hidden from the user to ensure transparent use. This feature is particularly important when interacting with older adults who may be reluctant to use new technologies~\cite{Brooks}.
The pen captures eight time series during handwriting: time stamps, 3-axis linear acceleration signals, 3-axis angular rates and the force applied on the pen tip. All signals are sampled at 50 Hz.

\subsection{Participants and protocol}
\label{protocol}
We recruited 80 healthy participants aged between 20 and 90 years. Any diagnosis of neurological, vascular or musculoskeletal disorders of the upper limbs was an exclusion criterion. Subjects over 65 years of age were included after verification of a Mini-Mental State Examination (MMSE)~\cite{Larner} score greater than 25.
All subjects wrote a free text (\emph{Text}, up to 10 lines) and a shopping list (\emph{List}, up to 8 words). The tasks had no specific constraints to make them very similar to everyday writing. The Ethics Committee of the Politecnico di Milano approved the study protocol (n. 10/2018).

\subsection{Calculation of handwriting and tremor indicators}

A set of 14 parameters related to handwriting kinematics and dynamics and to tremor were extracted from the raw data collected during each of the two writing tasks. The calculation was implemented in Matlab\textregistered\, R2020b (Mathworks\textregistered, Natick, MA USA)\footnote{See Lunardini et al.~\cite{lunardini_difebbo} for a detailed description of the indicators}. 
The following indicators were calculated:

\begin{itemize}

\item \emph{Temporal handwriting measures}. 
Starting from the writing force signal, we divided handwriting into strokes, defined as the writing segments where the pen tip was in contact with the paper surface (non-zero force tracts). We then considered the averaged stroke duration within a writing task as the mean on-sheet time ($OnSheet$). Similarly, we kept the averaged duration of the non-writing segments (zero-force tracts) as the mean in-air time ($InAir$). The in-air time intervals longer than 2 seconds were excluded as we treated them as pauses. The ratio of the latter to the former was defined as the air-sheet time ratio ($AirSheetR$). These temporal parameters have been shown to grow with users' age~\cite{SaraandEngel}.

\item \emph{Pen Tilt}. The tilt angle of the pen was calculated using the sensor fusion algorithm described in~\cite{lunardini_difebbo}. We retained the mean ($Tilt_{Mean}$), coefficient of variation ($Tilt_{CV}$) and variance ($Tilt_{Var}$) of the tilt angle signal during writing (pauses excluded). We considered an angle of 90$^{\circ}$ for the pen in vertical position. Previous studies have also included pen tilt to characterise handwriting in different conditions~\cite{Asselborn, Garre-Olmo}.

\item \emph{Writing Force}. Mean writing force ($Force$) was calculated by averaging the force signal over all strokes recorded during the writing task. The mean number of force changes ($NCF$), calculated as the average number of local maxima and minima within a stroke, was also retained as a measure of force variability. Force and force variability have been shown to change with age in handwriting~\cite{Drotar}.

\item \emph{Writing Smoothness}. We calculated the number of acceleration changes ($NCA$) as the average number of local minima and maxima in the 3D acceleration signal over all strokes. This quantity was observed to decrease with age~\cite{Walton}.
\end{itemize}

To extract tremor, we divided the linear acceleration recorded during a writing task into 500 sample segments~\cite{Meigal}. We computed the power spectrum for each segment using the Hilbert-Huang transform (HHT)~\cite{Huang}, which has been preferred in the literature for the study of voluntary tremor over the standard Fourier transform~\cite{zhang2008detection}. The following tremor indicators were then calculated:

\begin{itemize}
\item \emph{Tremor frequency}. We obtained the mean modal frequency ($F_{modal}$) by averaging the frequencies of the highest peak in the power spectrum over all the segments~\cite{Hong}.

\item \emph{Tremor Amplitude}. We calculated the root mean square (RMS) of the tremor signal and retained the mean $RMS$ by averaging the root mean square of the power spectrum over all segments.

\item \emph{Tremor entropy}. We considered the approximate entropy measure ($ApEn$), as in our previous study~\cite{lunardini_difebbo}. The entropy value (between 0 and 2) measures the unpredictability of the acceleration signals, which can be influenced by the higher or lower regularity of the tremor components. Entropy has been measured to decrease with age and pathology~\cite{Vaillancourt}.

\item \emph{Nonlinear characteristics of tremor}. We applied the recurrence quantification analysis (RQA) to the acceleration signals. As in~\cite{Meigal}, we retained the recurrence ratio ($RR$) to measure the tendency of the tremor dynamics to express repeated patterns in time and the percentage of determinism ($DET$) to estimate the predictability of the gestures during handwriting.
\end{itemize}

\subsection{Classification tasks}
Following the protocol described in Section \ref{protocol}, we defined $D_{T}$ the Text dataset and $D_{L}$ the List dataset, both consisting of 80 samples and 15 attributes (14 indicators and the group label). We also created the $D_{TL}$ dataset, consisting of 80 samples per 29 attributes (28 indicators of both tasks and the group label), by merging the two.
Given the set of four ordered age intervals $A=\{YY\in[20,40),EY\in[40,60),EF\in[60,70),EE\in[70,95)\}$ and the set of writing task data $W=\{T, L,TL\}$, we define $D_{w}^{a}$ with $a \in A$, $w \in W$ as the data set composed of the 20 samples and computed from the group $a$ over the task $w$. 


We investigated the ability of the handwriting features to discriminate between subjects of different age groups. The indicators measure high-level phenomena  underlying the complex handwriting process, which is strongly influenced by ageing. Therefore, we used machine learning  classification techniques to account for the multivariate and non-linear nature of the problem.

Two different machine learning algorithms were chosen to compare different classification logics. We used Logistic Regression as a baseline performance measure, as it is one of the simplest and most commonly used linear classifiers. The second was a more recent boosting algorithm called Catboost~\cite{dorogush2018catboost}. This algorithm is known to achieve remarkable performance while avoiding data overfitting, even with small datasets~\cite{prokhorenkova2017catboost}.

\begin{figure*}[t]
\centering
\includegraphics[width=16cm]{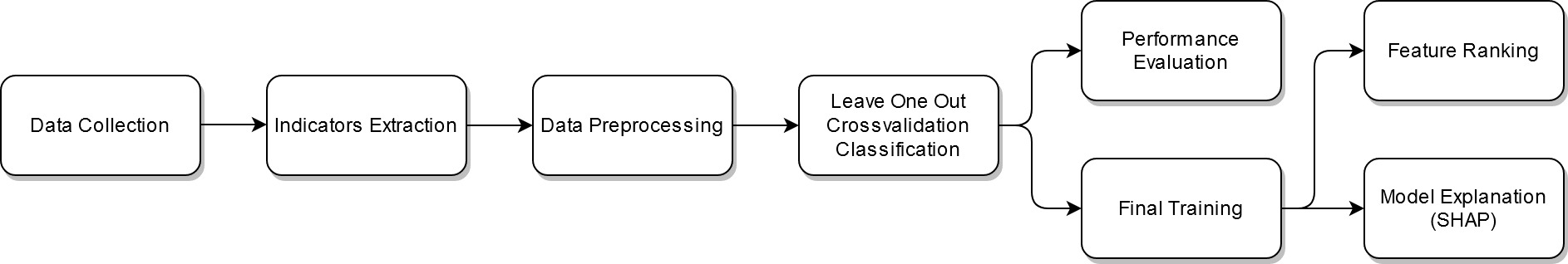}
\caption{The data processing, classification and model explanation workflow.}
\label{fig:pipeline}
\end{figure*}

Since the goal of our analysis is the detection of age-related anomalies in handwriting data, we focus on binary classification tasks to discriminate between age group differences. In detail, we computed two pools of classification tasks: the first one is between adjacent groups by age, i.e. $YY$vs$EY$, $EY$vs$EF$ and $EF$vs$EE$. The second is between $YY$, $EY$ and the $EE$ group. The first task pool is to evaluate the performance of the classifiers in discriminating between groups within adjacent age ranges. Models that perform well on these tasks are expected to be more sensitive to minimal changes in handwriting due to the age decline process. The second pool of tasks was designed to assess the greater ability of the models to detect more relevant changes in more distant age groups.

For each task we applied a data normalisation in the range [0,1]. The samples were then labelled 1 for the oldest group and 0 for the youngest. In this way, the machine learning algorithms learned to predict the probability of the sample belonging to the correct class.
For each experiment, we evaluated the models according to a wide range of classification metrics: Accuracy, Precision, Recall, F1 and Area Under the ROC Curve (ROC-AUC).
For monitoring purposes, Precision is the most important metric as it measures how robust the classifier is in determining the true positives.

To obtain a less biased performance estimate, we evaluated both Logistic Regression and Catboost with default parameters by Leave-One-Out (LOO) cross-validation with early stopping set to 20 epochs.
The full pipeline is shown in Fig. \ref{fig:pipeline}: after collecting the subjects' data and extracting the indicators, they are preprocessed and prepared to be learned by the proposed models. This learning phase involves the use of the LOO cross-validation technique mentioned above, which provides an estimate of the performance on unlearned data and the best number of learning iterations for each model evaluated. Finally, a model is trained on the entire dataset for a number of epochs equal to the average of those just found. This is done for the sole purpose of interpreting and ranking the most important features through model explanation techniques.

\subsection{Model explanation techniques}
We used a model explanation technique to overcome the limitations of the black-box nature of the Catboost classification algorithm and to obtain precise information about the model's decisions, i.e. the importance and role of the handwriting indicators in predicting the subject's age group. 
We used SHAP~\cite{lundberg2017unified, lundberg2020local2global}, a model explanation library based on game theory that computes the Shapley values~\cite{roth1988shapley} of the features according to their impact on its predictions.
In a binary classification task, SHAP first computes the baseline prediction value, i.e. the mean value predicted by the model given the observed samples, and then assigns a real number to weight each feature according to its average contribution in feature coalitions, i.e. its Shapley value.
It is then possible to explore the role of each feature in the classification of individual samples, independent of the fact that the model has learned them during the training step. The sample prediction represents the sum of the feature contribution starting from the baseline. If a feature has a positive influence, it influences the prediction in favour of class 1 and vice versa.
This step was useful to understand, for each sample and age group, how much each indicator leads the model to predict class 0 or 1.

\section{Results}\label{sec:results}

All participants were divided into four groups defined by age: group YY between 20 and 39 (12 males, 8 females, mean age 27.4$\pm$2.4), group EY between 40 and 59 (12 males, 8 females, mean age 57. 7$\pm$6.28), group EF between 60 and 69 (10 men, 10 women, mean age 65.45$\pm$2.2), and subjects older than 70 (6 men, 14 women, mean age 80.2$\pm$7) were included in group EE. Each group contained 20 subjects.
Tab. \ref{tab:logistic_table} and Tab. \ref{tab:catboost_table} report the performance metrics for each classification task and dataset for the Logistic Regression and the Catboost, classifier respectively. 

As expected, the Catboost algorithm performed best. 
The detailed results of 3 classification tasks are shown in Fig. \ref{fig:final_plot}: $EY$vs$EF$ in the first column of the figure, $EF$vs$EE$ in the second column and $EY$vs$EE$ in the third column. The first and second tasks involved the most interesting class for monitoring purposes (the EF, with individuals in the 60-69 age range) and its two closest classes in terms of age ranges (40-59 and 70+ respectively). The third task was instead designed to assess how much the age gap of 60-69 years improved the binary classification between the younger and older groups of individuals. 
Row (a) shows the ROC-AUC performance obtained with Catboost trained and evaluated on the Text, List and Text+List datasets. In all cases, the results of the Text dataset achieved the highest ROC-AUC. For these reasons, the plots in rows (b), (c) and (d) show the results of the Text dataset only. Row (b) shows the confusion matrices and 
Rows (c) and (d) show the final SHAP feature ranking models, trained on the full task datasets and tuned via LOO cross-validation. While row (c) shows the absolute influence of the features, row (d) shows the same ranking and explains how the learned samples were predicted according to their feature values. Each point in the figures in row (d) represents the Shapely value of the feature for a particular sample. The blue-red colour scale indicates the value of the indicator (low to high), and the negative Shapely values pushed the prediction towards class 0 (the youngest group), while the positive values favoured the classification of the subject in class 1 (the oldest group). The results of these three tasks are detailed in the following subsections.

\begin{table*}[tbp]
    \caption{Logistic regression scores evaluated by LOO cross-validation for age group binary classification. For each classification task, the best scores are highlighted in bold.}
    \label{tab:logistic_table}
    \centering
    \setlength{\tabcolsep}{1.15px}
    \begin{tabular}{|c|ccc|ccc|ccc|ccc|ccc|}
    \hline
    & \multicolumn{3}{c}{Accuracy} & \multicolumn{3}{c}{Precision} & \multicolumn{3}{c}{Recall}& \multicolumn{3}{c}{F1}& \multicolumn{3}{c|}{ROC-AUC}\\
                   & Text & List & Text+List & Text & List & Text+List & Text & List & Text+List & Text & List & Text+List & Text & List & Text+List\\
    
    YY vs EY & 57.5 & $\mathbf{65.0}$ & $\mathbf{65.0}$    & 58.8 & $\mathbf{68.8}$ & $\mathbf{68.8}$   & 50.0 & $\mathbf{55.0}$ & $\mathbf{55.0}$   & 54.1 & $\mathbf{61.1}$ & $\mathbf{61.1}$   & 56.5 & $\mathbf{68.2}$ & 68.0\\
    EY vs EF & $\mathbf{67.5}$ & 65.0 & 62.5    & $\mathbf{68.4}$ & 62.5 & 61.9   & 65.0 & $\mathbf{75.0}$ & 65.0   & 66.7 & $\mathbf{68.2}$ & 63.4   & 72.5 & 65.8 & $\mathbf{73.5}$\\
    EF vs EE   & 72.5 & $\mathbf{77.5}$ & 72.5    & 69.6 & $\mathbf{78.9}$ & 71.4   & $\mathbf{80.0}$ & 75.0 & 75.0   & 74.4 & $\mathbf{76.9}$ & 73.2   & 80.5 & 81.2 & $\mathbf{83.8}$\\
    YY vs EE   & 90.0 & $\mathbf{92.5}$ & $\mathbf{92.5}$    & 94.4 & 94.7 & $\mathbf{100.}$   & 85.0 & $\mathbf{90.0}$ & 85.0   & 89.5 & $\mathbf{92.3}$ & 91.9   & 93.0 & 94.2 & $\mathbf{98.2}$\\
    EY vs EE   & 85.0 & $\mathbf{87.5}$ & 85.0    & 88.9 & $\mathbf{89.5}$ & 85.0   & 80.0 & $\mathbf{85.0}$ & $\mathbf{85.0}$   & 84.2 & $\mathbf{87.2}$ & 85.0   & 88.8 & $\mathbf{95.8}$ & 95.0\\
    \hline
    \end{tabular}
\end{table*}

\begin{table*}[tbp]
    \caption{Catboost scores evaluated by LOO cross-validation for age group binary classification. For each classification task, the best scores are highlighted in bold.}
    \label{tab:catboost_table}
    \centering
    \setlength{\tabcolsep}{1.15px}
    \begin{tabular}{|c|ccc|ccc|ccc|ccc|ccc|}
    \hline
    & \multicolumn{3}{c}{Accuracy} & \multicolumn{3}{c}{Precision} & \multicolumn{3}{c}{Recall}& \multicolumn{3}{c}{F1}& \multicolumn{3}{c|}{ROC-AUC}\\
                   & Text & List & Text+List & Text & List & Text+List & Text & List & Text+List & Text & List & Text+List & Text & List & Text+List\\
    YY vs EY & 82.5 & $\mathbf{85.0}$ & 80.0    & $\mathbf{84.2}$ & 81.8 & 83.3   & 80.0 & $\mathbf{90.0}$ & 75.0   & 82.1 & $\mathbf{85.7}$ & 78.9   & 93.0 & $\mathbf{96.8}$ & 95.5\\
    EY vs EF & $\mathbf{90.0}$ & 82.5 & 85.0    & 83.3 & $\mathbf{84.2}$ & 81.8   & $\mathbf{100.}$ & 80.0 & 90.0   & $\mathbf{90.9}$ & 82.1 & 85.7   & $\mathbf{98.0}$ & 92.2 & 96.2\\
    EF vs EE   & $\mathbf{90.0}$ & $\mathbf{90.0}$ & 85.0    & $\mathbf{94.4}$ & $\mathbf{94.4}$ & 85.0   & $\mathbf{85.0}$ & $\mathbf{85.0}$ & 85.0   & $\mathbf{89.5}$ & $\mathbf{89.5}$ & 85.0   & $\mathbf{98.1}$ & 97.0 & 95.5\\
    YY vs EE   & $\mathbf{97.5}$ & 90.0 & $\mathbf{97.5}$    & 95.2 & 94.4 & $\mathbf{100.}$   & $\mathbf{100.}$ & 85.0 & 95.0   & $\mathbf{97.6}$ & 89.5 & 97.4   & 99.5 & 98.8 & $\mathbf{100.}$\\
    EY vs EE   & $\mathbf{92.5}$ & $\mathbf{92.5}$ & 87.5    & $\mathbf{94.7}$ & $\mathbf{94.7}$ & 89.5   & $\mathbf{90.0}$ & $\mathbf{90.0}$ & 85.0   & $\mathbf{92.3}$ & $\mathbf{92.3}$ & 87.2   & $\mathbf{99.5}$ & 98.4 & 98.2\\
    \hline
    \end{tabular}
\end{table*}

\subsection{EY vs EF}
In the $EY$ vs $EF$ task, the ROC curves show that the Catboost models trained on the Text data perform best with an AUC of 98.0\%. The corresponding confusion matrix for the text data shows that there are no false negatives, which translates into a recall of 100\%. In total there are 4 false positives.
According to the Shapley values of the final model, this task was strongly influenced by the $Tilt_{Mean}$, $ApEn$ and $Force$ indicators. In particular, high values of $Tilt_{Mean}$, $ApEn$ and $Force$ were associated with younger subjects belonging to age class $EY$.

\subsection{EF vs EE}
In the $EE$ vs $EF$ task, the ROC curves show that the Catboost models trained on the Text data achieve the best results with an AUC of 98.12\%, meaning that it is also possible to discriminate between the two oldest classes. 
The corresponding confusion matrix for the Text data shows the same behaviour as in the previous task, with only 1 false positive and 3 false negative predictions, resulting in an accuracy of 90\% and a precision of 94.4\%.
According to the Shapley values of the final model, the $InAir$ indicator proved to be the core feature for this binary classification. The model strongly related high values of the $InAir$ indicator to the oldest class $EE$.

\subsection{EY vs EE}
Finally, in the $EY$ vs $EE$ task, the ROC curves show that the Catboost model trained on the text data performs best with an AUC of 99.5\%, meaning that the predictions of the unseen data are almost perfectly accurate. The corresponding confusion matrix shows only 1 false positive and 2 false negatives, resulting in a balanced F1 score of 92.3\%. According to the Shapley scores of the final model, this task was heavily influenced by $InAir$ and $ApEn$, and less so by $F_{modal}$ and $DET$. The high values of $InAir$ and $DET$ pushed predictions in favour of the oldest class $EE$, while high values of $ApEn$ and $F_{modal}$ were more associated with the youngest subjects belonging to class $EY$.

\begin{figure*}[!tbp]
\begin{center}
\setlength{\tabcolsep}{3.5px}
\begin{tabular}{cccc}
\multicolumn{4}{c}{}\\
& \large{EY vs EF} & \large{EF vs EE} & \large{EY vs EE}\\

\large{a)} & \raisebox{-.5\height}{\includegraphics[width=0.305\textwidth]{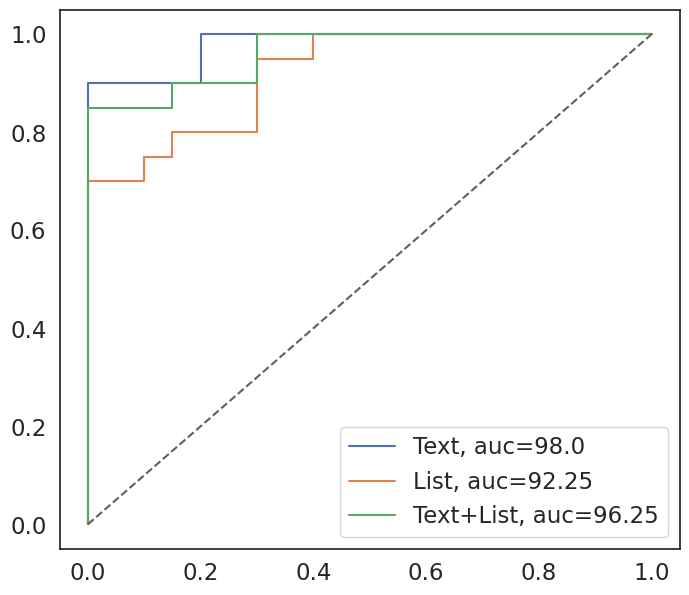}} & \raisebox{-.5\height}{\includegraphics[width=0.305\textwidth]{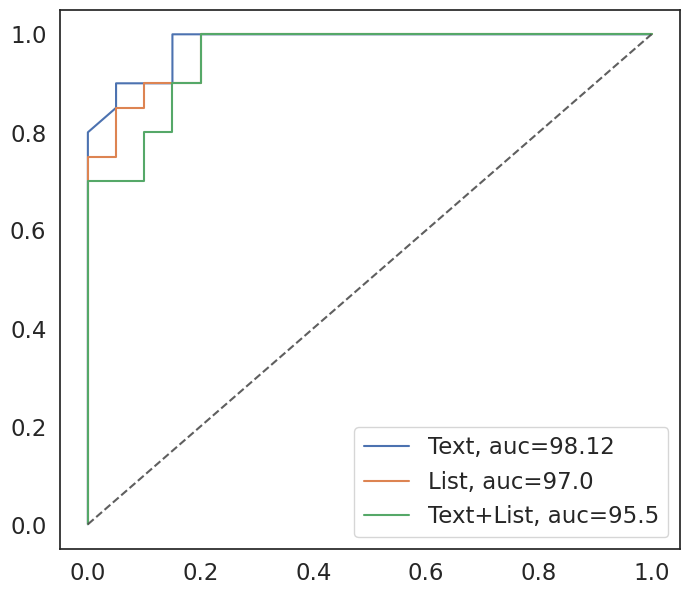}} &
\raisebox{-.5\height}{\includegraphics[width=0.305\textwidth]{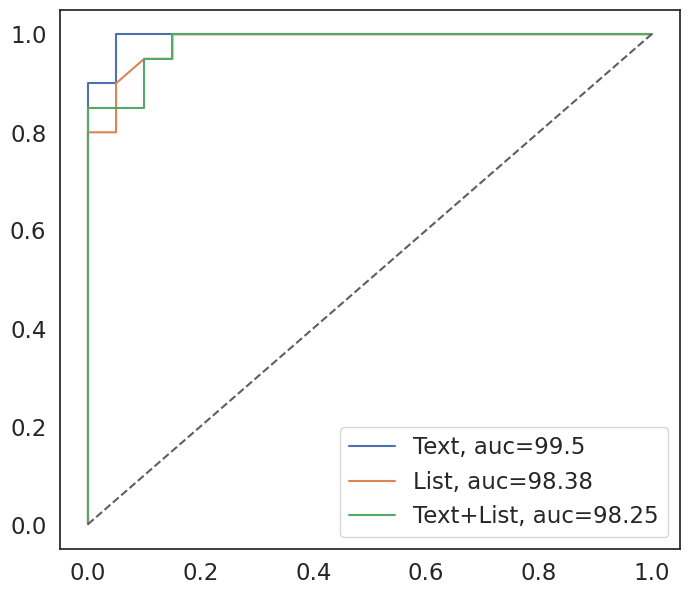}}\\

\large{b)} & \raisebox{-.5\height}{\includegraphics[width=0.305\textwidth]{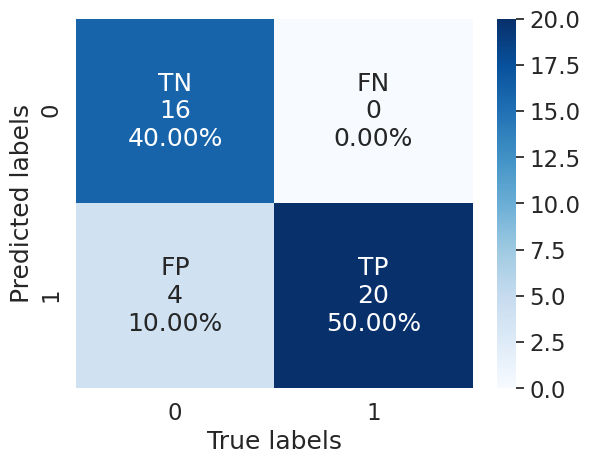}} & \raisebox{-.5\height}{\includegraphics[width=0.305\textwidth]{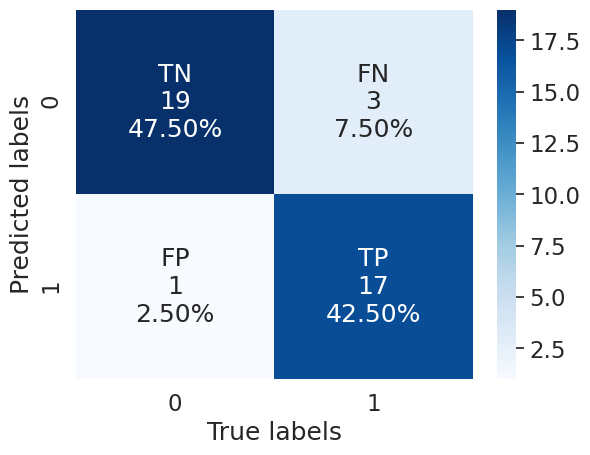}} &
\raisebox{-.5\height}{\includegraphics[width=0.305\textwidth]{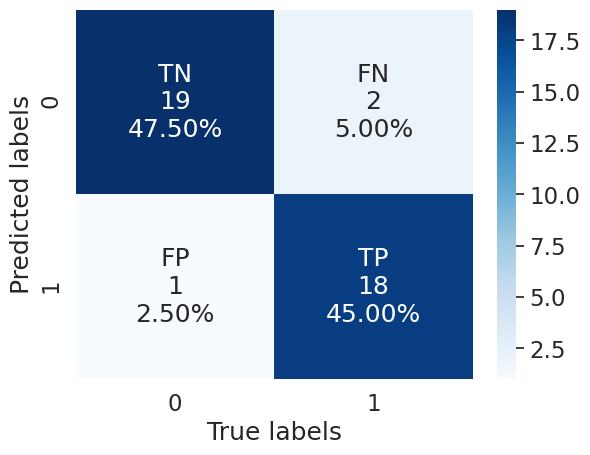}}\\

\large{c)} & \raisebox{-.5\height}{\includegraphics[width=0.305\textwidth]{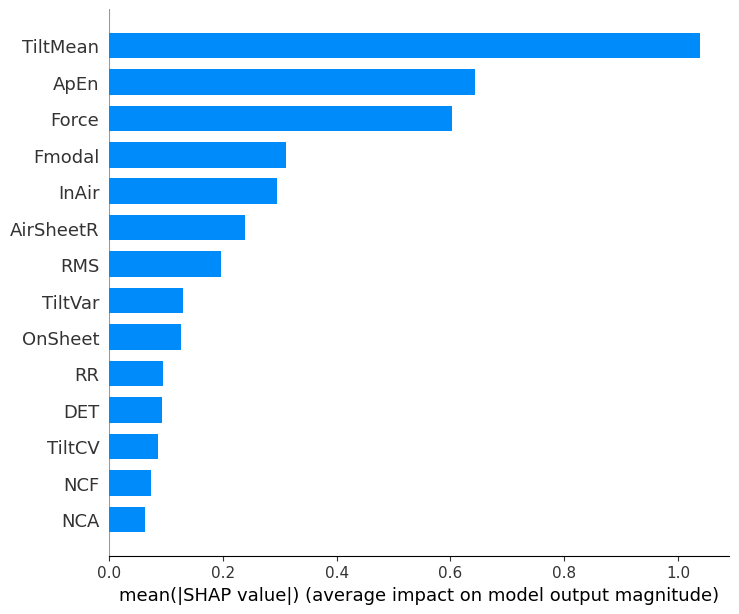}} & \raisebox{-.5\height}{\includegraphics[width=0.305\textwidth]{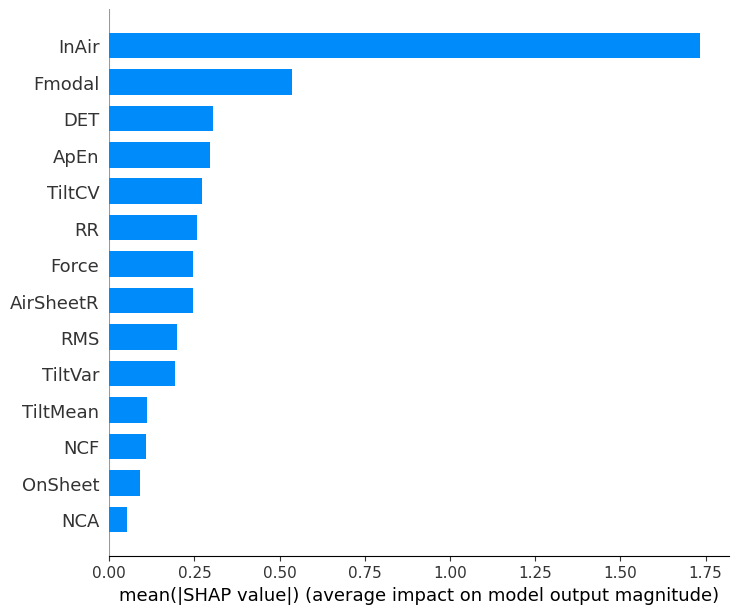}} &
\raisebox{-.5\height}{\includegraphics[width=0.305\textwidth]{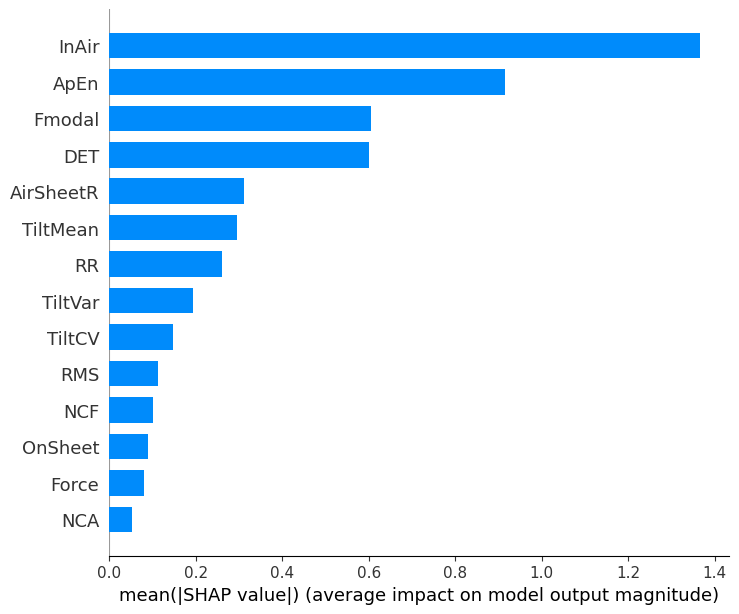}}\\

\large{d)} & \raisebox{-.5\height}{\includegraphics[width=0.305\textwidth]{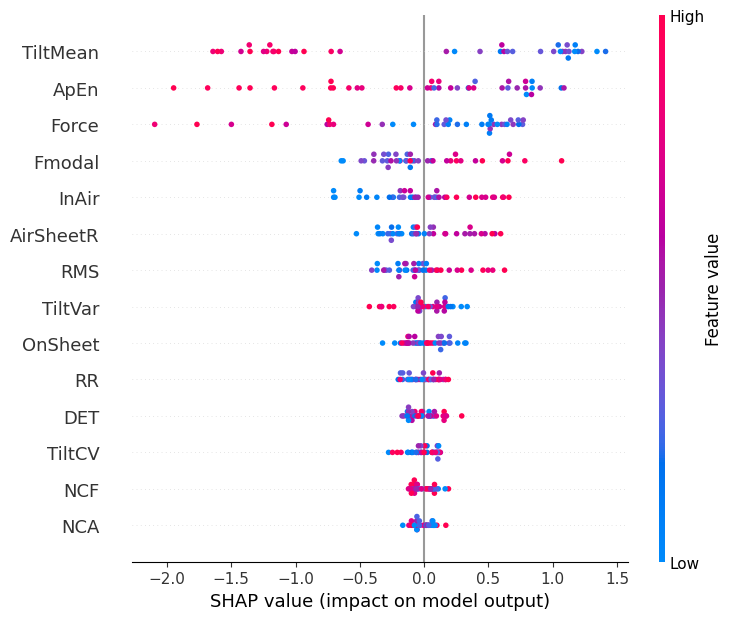}} & \raisebox{-.5\height}{\includegraphics[width=0.305\textwidth]{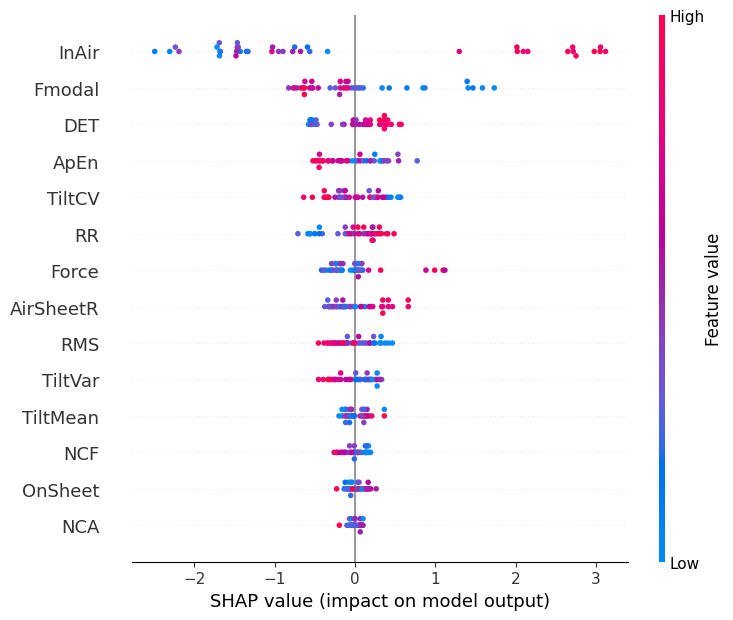}} &
\raisebox{-.5\height}{\includegraphics[width=0.305\textwidth]{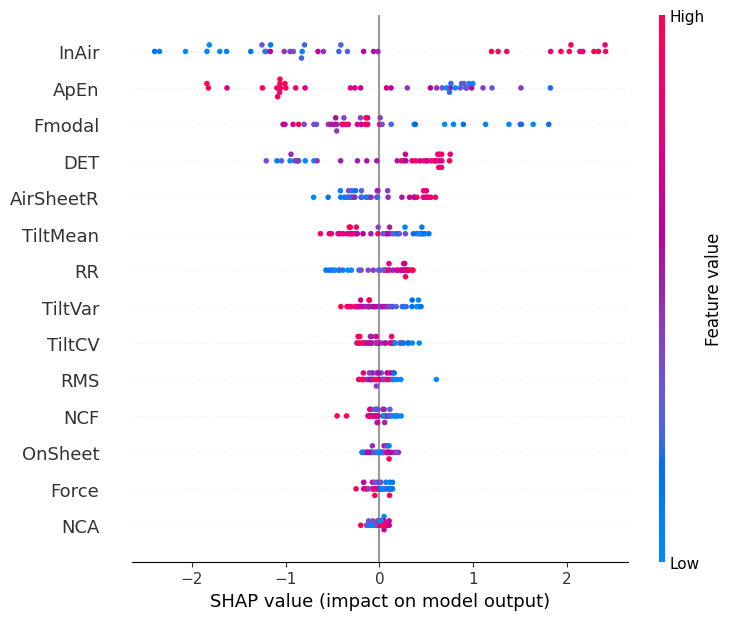}}\\

\end{tabular}
\caption{Classification performance and model explanation plots for the EY vs EF, EF vs EE and EY vs EE tasks: the ROC-AUC metrics achieved by the Text, List and Text+List indicators are in row a); the confusion matrices are in row b); rows c) and d) report the absolute average Shapely values and the Shapely value of the features for each sample, respectively. Group EY includes subjects aged between 40 and 59 years, group EF includes subjects aged between 60 and 69 years, and group EE includes subjects aged over 70 years.}
\label{fig:final_plot}
\end{center}
\end{figure*}

\section{Discussion}\label{sec:discussion}

In this paper we have demonstrated the utility of quantitative analysis of handwriting in discriminating between healthy subjects of different age groups. We used a novel smart ink pen to collect handwriting data during tasks that mimicked everyday writing. In fact, participants were asked to write a short free text and a shopping list without any constraints on content or writing modality. This particular setting was chosen to maximise ecological validity, with the ultimate goal of using our findings to develop, in the future, home-based solutions dedicated to the early detection of decline in seniors. Therefore, particular attention was paid to the correct classification of the individual's age, as the association of their handwriting characteristics with an older age group could be interpreted as a clinically relevant anomaly~\cite{Rosales}.

One of the most powerful and advanced machine learning classification algorithms, i.e. Catboost, and a more traditional one, i.e. Logistic Regression, were used to carry on four different binary classification tasks, based on the set of handwriting indicators we computed from the raw free text and shopping list data. Our results showed that the Catboost algorithm outperformed Logistic Regression in almost all the tasks and datasets we considered. The improvements were sensibly more significant in the classifications between groups with close age ranges (the first pool of tasks), where the differences in individuals' handwriting were expected to be smaller. This confirmed Catboost's superior sensitivity to changes in handwriting indicators in terms of a baseline estimator.

In the first pool of tasks, we considered the inter-group classifications with individuals in close age ranges (i.e. $YY$vs$EY$, $EY$vs$EF$ and $EF$vs$EE$). The aim of these tasks was to test the sensitivity of the models to small variations in handwriting performance that might be expected between healthy individuals with small age differences~\cite{Spaan}. Very good to excellent performances (accuracy between 82.5\% and 90\%, precision from 81.8\% to 94.4\%, recall from 75\% to 100\% and ROC-AUC from 92.2\% to 99.5\%) were obtained in the classification of the first pool, considering all three datasets composed by the indicators calculated from the Text, List and combined Text-List data. These results showed the good ability of the models to detect slight handwriting variations in healthy subjects. Therefore, we could expect a high sensitivity to the changes in handwriting data due to an abnormal or pathological ageing decline~\cite{Spaan}.

In the second pool of tasks, we looked instead at classifications between more distant age groups. As expected, scores were generally higher on these tasks, as the differences in handwriting should have been more pronounced. In the classification between $YY$-$EE$, i.e. the more distant classes, the best accuracy (97.5\%) was achieved using only the Text indicators and the combined set of Text and List indicators. Perfect precision and ROC-AUC (100\%) were obtained using the Text+List data and perfect recall using the Text data alone. The last setting, $EY$vs$EE$, showed high evaluation metrics with both Text and List data, all over 92.3\%.

For the classification between the younger groups, $YY$ and $EY$, i.e. 20-39 and 40-59 years of age, the list turned out to be the data on which the model performed better in terms of accuracy, recall, F1 and ROC-AUC. For the other adjacent groups of tasks, $EY$-$EF$ and $EF$-$EE$ (i.e. 40-59 vs. 60-69 years and 60-69 vs. 70+ years), the best performance was obtained with the text in almost all cases. This slight task dependency of the classification results might be related to some differences between the Text and List tests. Also, the writing dynamics might have been partly influenced by the type of task, since each item in the List was written in a new line and with single words, as articles and conjunctions were less frequent if not absent. In addition, writing a free text generally required a greater cognitive effort, which might explain the higher classification performances in  $EY$-$EF$ and $EF$-$EE$ . Nevertheless, the differences in the results were small and, given the relatively small number of samples, chance factors could not be excluded. Further research is planned. However, the results suggest that both data collection methods are still valid and contain intrinsic age-related information.

We further analysed our experiments using the SHAP model explanation technique. It was useful to understand the impact of each handwriting indicator in the different tasks and to see their behaviour. In this paper we detailed the results and analysis of three classifications: the first two belonged to the first pool and included the class of individuals in the range 60-69 years ($EF$), which is the more critical for the purpose of early detection of decline; the third consisted in the classification between individuals in the age ranges 40-59 and 70+, and it aimed at showing the more marked difference in handwriting characteristics between the two classes.

In the first classification, the groups aged 40 to 59 and 60 to 69 ($EY$ and $EF$) were considered. These two groups represent, respectively, a population of healthy subjects in which the effects of age decline should be absent, and a population in which a decline in physical or cognitive functionality may be at an early stage~\cite{Trevis, Gale}. As shown in Fig. \ref{fig:final_plot} row (b), the handwriting indicators were able to correctly classify all individuals in the 60-69 age range (with a recall score of 100\%), while four subjects in the 40-59 age range were misclassified (precision score equal to 84.2\%). Our results confirm previous findings in the literature, where it has been observed that handwriting varies significantly in middle, younger and older adults~\cite{Marzinotto}. According to Walton~\cite{Walton}, handwriting characteristics can be stable in healthy subjects for at least 5 years. In fact, the four false positives were all over the age of 52. Two of them were over 55. Therefore, it was likely that their handwriting characteristics results were closer to the older group.
The model explanation (Fig. \ref{fig:final_plot}, lines c and d) showed that both handwriting dynamics and tremor features were among the more influential.  

The tilt of the pen ($Tilt_{Mean}$) was the most important feature in the $EY$ vs $EF$ classification. According to Marzinotto et al.~\cite{Marzinotto}, a higher pen tilt (on the right) is typical in middle-aged adults ($EF$). The approximate entropy ($ApEn$) also played a significant role, indicating a lower predictability of the handwriting time series of the younger class. This result is in line with the findings of our previous work~\cite{lunardini_difebbo}, where, using similar experimental settings, significant differences between age groups were found. The trend of decreasing entropy with age was consistent with previous literature studying resting and postural tremor in younger and older adults~\cite{Sturman, Hong, Vaillancourt}. Although its variation was not statistically significant between different age groups in Lunardini et al.~\cite{lunardini_difebbo}, in the current study writing force ($Force$) emerged as the third most predictive feature in the $EY$-$EF$ classification. The predictions were shifted towards the older group ($EF$) when the force values were lower. This was in line with the study by Engel Yeger et al.~\cite{Engel-Yeger} in 2012, Caligiuri~\cite{Caligiuri2} in 2014 and Marzinotto et al.~\cite{Marzinotto} in 2016. 

In the following four features, sorted by decreasing importance, we found two frequency domain and two temporal parameters. The modal frequency had no significance in the statistical group differences in our previous work~\cite{lunardini_difebbo}, but it affects the classification $EY$ vs $EF$, linking higher values to the older class. The same behaviour was found for $RMS$. Previous studies in the literature show that some neurological conditions, such as Parkinson's disease, could affect the modal frequency~\cite{Suilleabhain}, while no apparent age effect on this parameter has been shown. The effect of the temporal indicators ($InAir$ and $AirSheetR$) was considerable, confirming the tendency of the older class to have more prolonged non-writing moments, found in our previous work~\cite{lunardini_difebbo}, and others~\cite{SaraandEngel, perla}.

The second classification, between the groups aged 60 to 69 and 70+ ($EF$ and $EE$), was the most relevant for investigating the suitability of our approach in the scenario of early detection of decline. In a normal ageing process, physical or cognitive decline is expected to be more consistent in the older group of people aged 70+~\cite{Trevis, Gale}. Therefore, whenever an individual in the younger group (aged 60-69) is associated with the older group, it could be interpreted as a sign of abnormal decline. In this task, the handwriting indicators were used to discriminate individuals in $EF$ from those in $EE$ with high performance scores. Our results showed that the $EF$-$EE$ classifier may be suitable for use in the monitoring of decline due to its high precision of 94.4\%. Only 1 subject out of 20 was wrongly classified as older, while the false negatives were 3 (Fig. \ref{fig:final_plot}, row (b)). The model explanation (Fig. \ref{fig:final_plot}, row (c) and (d)) showed that the in-air time parameter ($InAir$) was much more influential in the classification than all the others. As for the other tasks, higher $InAir$ were associated with individuals of the older class. Modal frequency was the second indicator of importance. The other indicators had quite similar effects, with frequency and non-linear features in higher positions. The tilt of the pen ($Tilt_{Mean}$) ended up among the last important indicators, although its variation ($Tilt_{cv}$) resulted in having a more significant impact. Nevertheless, all the indicators retained the same behaviour as in the previous tasks, thus confirming the consistency of the variations in handwriting measures with age.  

The third classification was between the $EY$ and the $EE$ groups, with individuals in the ranges 40-59 and 70+ years of age. The level of decline was expected to be very different among the healthy subjects' populations included in this task. As a consequence, the ability of the model in discriminating between these classes of individuals using the handwriting indicators resulted indeed increased. The Accuracy score was equal to 92.5\%, and the Precision was notably higher, with 94.7\%, at the expense of a minor Recall, equal to 90\%, with respect to the previous task. In fact, only one subject in $EF$ and two in $EE$ were wrongly classified. The model explanation (Fig. \ref{fig:final_plot}, rows (c) and (d)) showed almost the same indicators among the most relevant, however, some meaningful differences appeared. The $Tilt_{Mean}$ dropped from the first position in the classification $EY$-$EF$ to the sixth position in $EY$-$EE$ in the impact ranking while still keeping the same behaviour. In this task, $InAir$ emerged once again to have the highest impact,  with the same trend showing higher values in the older groups. With respect to the first classification task, the writing force dropped from the third position to the penultimate position, while determinism $DET$ raised to the fourth place, with the same impact of $F_{modal}$. Determinism was likely to increase with age as the influence of the predictable tremor components became more persistent.
All the handwriting indicators showed the same behaviour, in terms of value distribution, in the classifications of $EY$ with $EF$ and $EE$. Significant changes were found in the impact level of the tremor features, which resulted in more determinant in the discrimination between the two more distant groups, $EY$ and $EE$. 

The model explanation revealed that the impact of the handwriting indicators was task dependent, i.e., it changed according to the age ranges we considered in the classifications.
These differences in the feature importance highlighted the complexity of the age-driven decline in handwriting as the sensitivity of some indicators showed age-dependency. 
However, the behaviour of the indicators in the different age intervals was consistent with the previous findings in literature in populations of healthy subjects. This result reinforced the interpretation of the models, giving the possibility to understand their decisions as they relied to known handwriting-related quantities.

\section{Conclusion}\label{sec:conclusion}

In conclusion, this work showed the quantitative analysis of handwriting to classify individuals belonging to different age groups. Age-classifiers with high precision score may offer a novel and non-invasive instrument for the domestic monitoring of handwriting in elderly and frail individuals.
Our findings interest is enhanced by the innovative data acquisition modality we used to collect the subject's writing data, allowing the ecological assessment of daily-life handwriting. 

Age differences can be used to detect anomalies in handwriting, which may indicate an abnormal decline in individuals with the risk of developing pathological conditions. Moreover, more precise information about the nature of the conditions could be achieved by investigating more pathological-related handwriting changes, as Parkinson's disease and dementia, and developing illness-specific classifiers.

\section*{Acknowledgment}\label{sec:ack}
This work was supported by the European projects MOVECARE (Grant Agreement: 732158) and ESSENCE (Grant Agreement: 101016112).

\bibliographystyle{unsrt}  
\bibliography{references}

\end{document}